%% file: main.tex
\documentclass[runningheads]{llncs}

\usepackage{eccv}

\usepackage{eccvabbrv}

\usepackage{graphicx}
\usepackage{booktabs}

\usepackage[accsupp]{axessibility}  

\usepackage{multirow}
\usepackage{makecell}
\usepackage[table]{xcolor}
\definecolor{best}{RGB}{191, 255, 191}  
\definecolor{second}{RGB}{255, 255, 153} 
\definecolor{third}{RGB}{255, 195, 113}   

\usepackage{hyperref}

\usepackage{orcidlink}
\usepackage{marvosym}
\renewcommand{\thefootnote}{\textrm{\Letter}}
\begin{document}

\title{SplitAvatar: One-shot Head Avatar with Autoregressive Gaussian Splitting} 

\titlerunning{SplitAvatar: One-shot Head Avatar with Autoregressive Gaussian Splitting}

\author{Hongzhe Liao\inst{1} \quad
Chuhua Xian\inst{1}$^{\textrm{\Letter}}$ \quad
Hongmin Cai\inst{1} \\
Haiyang Liu\inst{2} \quad
Fa-Ting Hong\inst{3}$^{\textrm{\Letter}}$
}

\authorrunning{H.~Liao et al.}

\institute{South China University of Technology \and
University of Tokyo \and
The Hong Kong University of Science and Technology
}

\maketitle
\let\thefootnote\relax\footnotetext{$^{\textrm{\Letter}}$~Corresponding authors.}
\begin{center}
    \centering
    \captionsetup{type=figure}
    \includegraphics[width=0.98\textwidth]{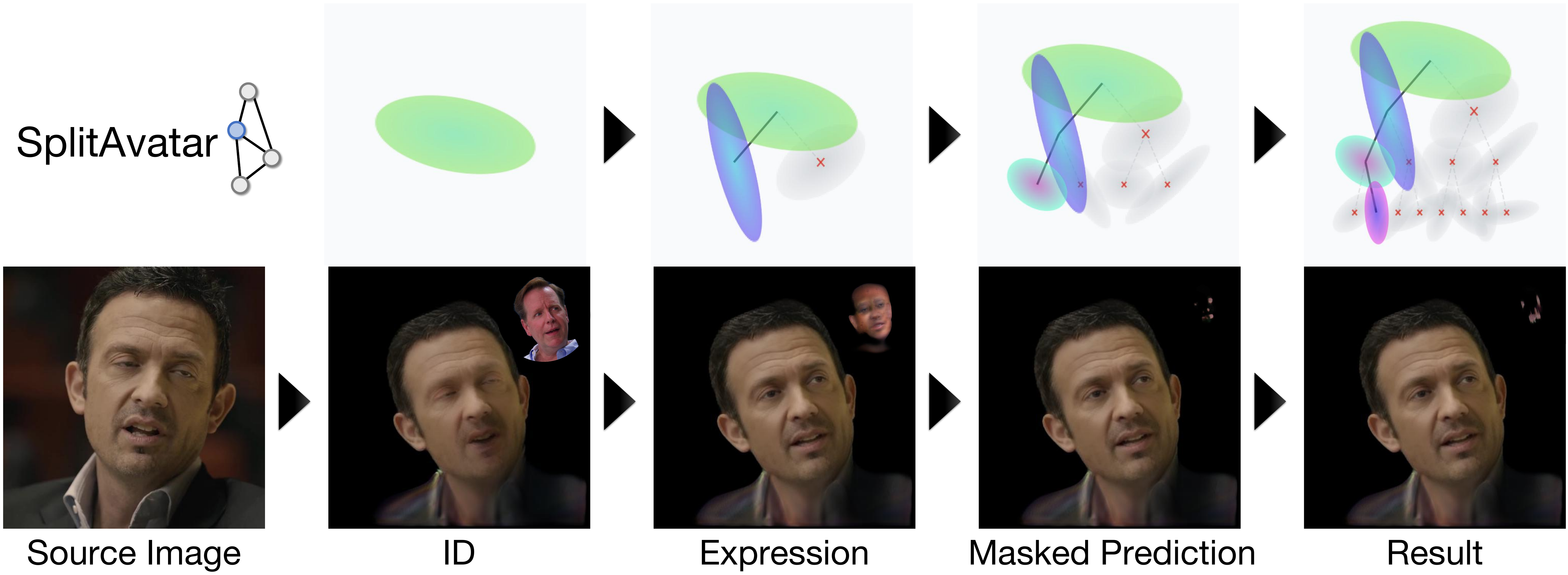}
    \captionof{figure}{SplitAvatar achieves high-quality one-shot head avatar synthesis by generating identity Gaussians and expression Gaussians separately. Expression details are then progressively refined from coarse to fine using an autoregressive architecture combined with a masked Gaussian split network.
    }
    \label{fig:teaser}
\end{center}

\makeatletter
\renewcommand{\fps@figure}{!tp}
\renewcommand{\fps@table}{!tp}
\makeatother

\input{sec/0_abstract}    
\input{sec/1_intro}

\input{sec/2_related_work}    
\input{sec/3_method}

\input{sec/4_experiment}

\input{sec/5_conclusion}


%
%
\bibliographystyle{splncs04}
\bibliography{main}

\clearpage
\appendix
\renewcommand{\thesection}{\Alph{section}}
\input{sec/6_suppl}

\end{document}

%% file: sec/0_abstract.tex
\begin{abstract}

3D Gaussian Splatting (3DGS) provides an efficient method for high-quality scene reconstruction using anisotropic Gaussians. Recently, 3DGS-based methods have significantly improved the rendering quality of human avatars while enabling real-time performance. However, existing methods suffer from a magnitude mismatch in the number of Gaussians generated by image-based and 3DMM-based approaches. This discrepancy results in reconstructed expressions that lack fine-grained detail. 
In this paper, we introduce a novel method for reconstructing an animatable head avatar from a single image. We propose a Graph splitting network to progressively generate Gaussians from coarse to fine using an autoregressive architecture. To address the graph inconsistency caused by split Gaussians, we employ a mesh topology extension method to align the GNN's connectivity with the increased Gaussian count. 
Furthermore, we introduce a novel density control method that includes a gating mechanism that generates soft masks for Gaussians, preventing over-densification after the splitting operation. This allows for dynamic control over Gaussian density across different facial regions. For smooth and rapid training, we employ a delayed filtering strategy to avoid re-computing the graph topology during training.
Experimental results demonstrate that our autoregressive structure effectively improves expression representation ability by progressively splitting Gaussians. This process, enabled by the GNN-guided splitting, synthesizes more precise facial details and achieves higher reconstruction quality.
\keywords{3D Gaussian splatting \and Avatar reconstruction \and Autoregressive}
\end{abstract}


%% file: sec/1_intro.tex
\vspace{-5pt}
\section{Introduction}
\label{sec:intro}
\vspace{-5pt}
The generation of one-shot human head avatars—animating a single source image using a driving video or image—is a critical task for applications like virtual digital humans and immersive online meetings. The objective is to synthesize a high-quality talking head that preserves the source identity while adopting the pose and expression of the driving input. The fundamental challenge of this task is the ambiguity: how to disentangle and reconstruct complex 3D facial geometry and appearance from a single 2D image.

To address this challenge, methods are broadly categorized as 2D-based or 3D-based. 2D-based methods~\cite{ren2021pirenderer,yin2022styleheat} warp source image features or pixels using an estimated 2D motion field and rely on a generative network to inpaint occluded regions and synthesize new content. While effective for simple scenarios, they inherently struggle with multi-view consistency and often produce visible artifacts under large pose variations. 3D-based methods overcome this limitation by explicitly modeling the geometric representation of the head avatar, yielding higher reconstruction quality and stronger multi-view consistency. These methods generally combine visual features with facial priors to synthesize head avatars. 3D Morphable Models (3DMMs), such as FLAME~\cite{li2017flame}, are widely adopted to represent digital human portraits and have proven to be an efficient and effective prior for constraining the reconstruction problem. Following advances in neural rendering, Neural Radiance Fields (NeRF)~\cite{mildenhall2021nerf} have been extensively explored and produced state-of-the-art results in talking head synthesis quality~\cite{li2023generalizable,ye2024real3d,chu2024gpavatar}. Despite their impressive quality, NeRF-based approaches rely on the volume rendering process, which incurs significant computational overhead and is prohibitive for interactive applications. Consequently, 3D Gaussian Splatting (3DGS)~\cite{kerbl20233d} and its variants have recently emerged as a powerful alternative, offering a promising path toward real-time, high-fidelity talking head synthesis.

However, existing 3DGS-based methods still face two significant challenges. First, they struggle with a fundamental scale mismatch between the facial prior and the Gaussian representation. Methods driven by 3DMMs, such as FLAME~\cite{li2017flame}, are constrained by a fixed number of vertices and a sparse edge topology, which inherently limits the scale of the reconstructed 3D Gaussians~\cite{chu2024generalizable}. This sparsity is difficult to reconcile with the dense 3D Gaussian representations required for high-fidelity head reconstruction. While some approaches attempt to generate new Gaussians via density control strategies~\cite{qian20243dgs} or by using mesh vertices as anchors to derive additional Gaussians~\cite{wei2025graphavatar}, they struggle to match Gaussian scenes at arbitrary scales and fail to robustly fuse the sparse expression Gaussians with the dense head Gaussians. Second, hierarchical or coarse-to-fine synthesis methods~\cite{hyun2024gsgan}, which attempt to build the Gaussian representation through a multi-level architecture, suffer from uncontrolled densification. These methods typically employ fixed replication rules that lack active filtering for the generated Gaussians. This leads to rampant redundancy and excessive Gaussian density, while simultaneously failing to allocate sufficient detail to critical expressive regions, such as the eyes and teeth.

To address these limitations, we propose SplitAvatar, a novel method for one-shot head avatar reconstruction. 
As illustrated in ~\cref{fig:struct}, SplitAvatar consists of two branches: one for avatar identity reconstruction and the other for expression driving. To capture the facial expression, we used the FLAME parameters obtained by 3DMM estimator to predict initial Gaussians. To increase the number of initial Gaussians, we then employ a Graph Neural Network (GNN) to split these Gaussians. This process conveys facial feature information between adjacent regions, with the graph structure built upon the topology of the FLAME model. To enable flexible control over the number of Gaussian splits and capture multiscale details, we adopt an autoregressive structure that simultaneously splits both the Gaussians and graph topology. We also apply a filtering module that learns a soft mask to determine which split Gaussians should be retained. This prevents over-densification and potential redundancy after splitting. Our method effectively resolves the magnitude difference in Gaussian point counts between expression transfer and head avatar reconstruction. SplitAvatar is highly configurable, allowing the number of generated Gaussians to be adjusted by tuning the parameters of the autoregressive structure. This enables adaptation to scenes of different complexity or seamless integration with other modules.

Our contribution can be summarized as follows:
\begin{itemize}
    \item We propose SplitAvatar, a novel framework for driving expressions from a single source image. It effectively fuses 3DMM-driven Gaussians with image-reconstructed Gaussians and resolves the magnitude mismatch between the two representations.
    \item We introduce an autoregressive GNN module to split the expression Gaussians, along with a graph extension rule to accommodate changes in the GNN topology. Furthermore, we utilize a Gaussian gating mechanism and a corresponding training strategy to dynamically control Gaussian density.
    \item Our method achieves state-of-the-art performance in one-shot head avatar reconstruction and expression transfer. Experiments and ablation studies validate the effectiveness of our proposed method to split Guassians to reproduce the portrait's identity and facial expression details.
\end{itemize}

%% file: sec/2_related_work.tex
\vspace{-5pt}
\section{Related Work}
\label{sec:related_work}
\vspace{-5pt}
\noindent\textbf{3D head avatar reconstructing.}
To better reconstruct 3D head avatars, early approaches utilized mesh reconstruction based on 3DMM~\cite{xu2020deep,khakhulin2022realistic}. With the rise of neural radiance fields (NeRF)~\cite{mildenhall2021nerf}, NeRF-based methods began to emerge, yet suffer from speed limitations, which is difficult to apply in real-time applications. Some studies attempted to generate portraits using 3DGS~\cite{kerbl20233d}, achieving real-time processing and promising results. Nevertheless, most of these require learning feature videos, which limits their practical applicability~\cite{hu2024gaussianavatar,xu2024gaussian,li2024talkinggaussian,ma20243d,wang2025mega}. CVTHead~\cite{ma2024cvthead} synthesizes controllable head avatars by utilizing point-based neural rendering and a vertex-feature transformer. Portrait4D~\cite{deng2024portrait4d} learns a transformer-based animatable tri-plane on synthetic multi-view datasets. Portrait4D-v2~\cite{deng2024portrait4d2} improves 4D consistency by training on pseudo multi-view videos generated from monocular input. GPAvatar~\cite{chu2024gpavatar} reconstructs generalizable head avatars by integrating a multi-triplane representation with a dynamic, point-based expression field. GAGavatar~\cite{chu2024generalizable} adopted a dual-lifting method to generate Gaussians from a single image. LAM~\cite{he2025lam} presents a large-scale feed-forward model that generates one-shot animatable Gaussian head avatars, using a transformer to predict FLAME-driven 3D Gaussians from a single image in seconds.

\begin{figure*}[t]
    \centering
    \includegraphics[width=1\linewidth]{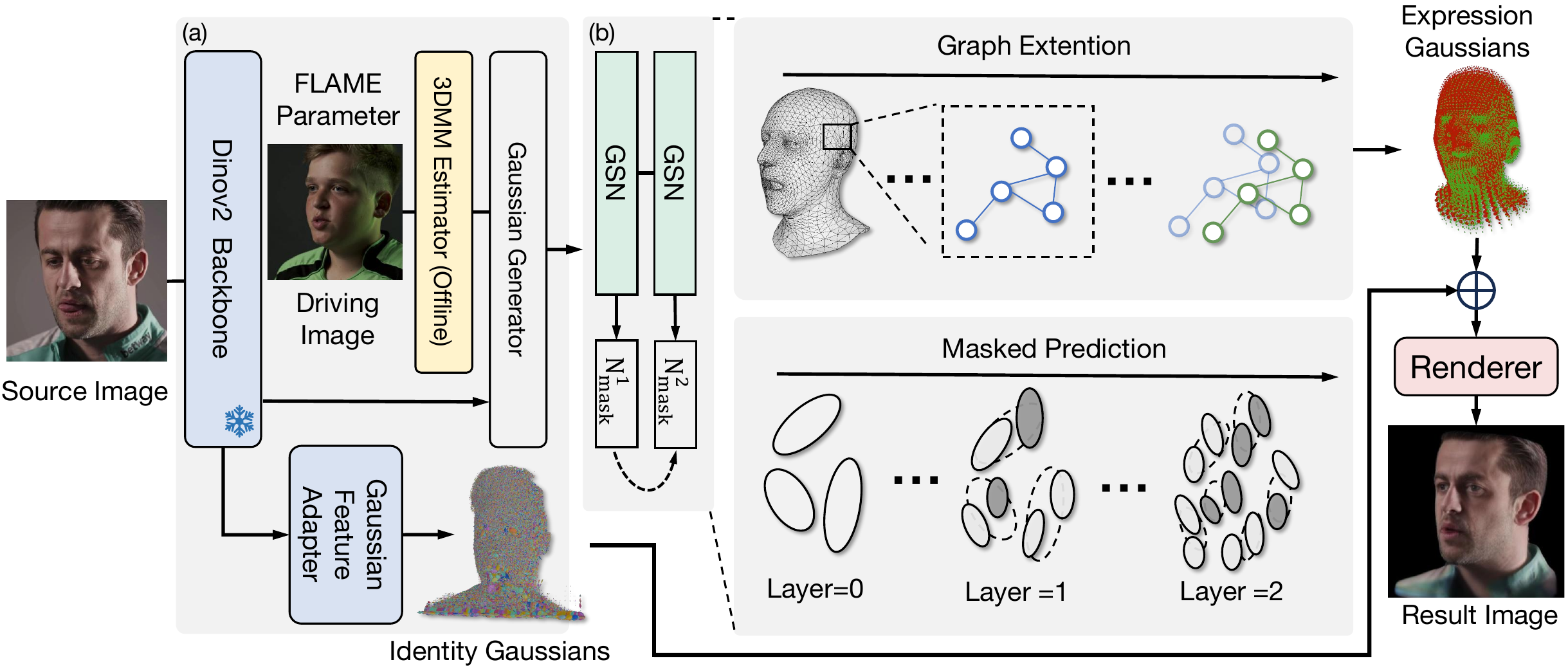}
    \vspace{-20pt}
    \caption{Overview of \textbf{SplitAvatar.} (a) we use a gaussian generator to predict initial expression Gaussians using the 3DMM estimator and identity features. We utilize a Gaussian feature adapter to predict identity Gaussians by visual features extracted from Dinov2. (b) the autoregressive graph splitting network predicts the next layer of expression Gaussians based on the previous layer. (\cref{subsec:GSN}) We also apply a soft mask network at each layer to filter out redundant split Gaussians. (\cref{subsec:mask})}
    \label{fig:struct}
    \vspace{-10pt}
\end{figure*}

Separately, several works have combined Graph Neural Networks (GNNs) with 3DGS. Zhang \etal~\cite{zhang2024gaussian} proposed the Gaussian Graph Network, which constructs Gaussian Graphs to model relationships between Gaussian groups from different views, yet it does not utilize graph information from 3DMM meshes. GraphAvatar~\cite{wei2025graphavatar} trains a geometric GNN and an appearance GNN to generate 3D Gaussian attributes from a tracked 3DMM, with the aim of reducing storage consumption. However, it requires per-identity training and lacks generalization capability. In contrast, our method attempts to create flexible number of Gaussians by a splitting strategy, thereby achieving a generalizable 3DGS facial representation while leveraging GNNs to capture facial information from 3DMMs.

\noindent\textbf{Autoregressive structure.}
Autoregressive models are fundamental to sequence data modeling and processing. 
With the advancement of deep learning, autoregressive models have been applied to model complex, high-dimensional data, such as text and images. In the field of computer vision, early attempts generated images in a raster-scan manner, either directly or in the latent space~\cite{van2017vq-vae,razavi2019vq-vae-2,chen2020generative,esser2021vqgan}. Visual Autoregressive Modeling~\cite{tian2024var} adopts a multi-scale structure with a coarse-to-fine, next-scale prediction process to generate high-resolution images, moving beyond traditional per-pixel or per-token prediction methods. Kumbong \etal~\cite{Kumbong_2025_hmar} combine next-scale prediction with masked prediction~\cite{chang2022maskgit,chang2023muse,li2023mage}, further improving the model's generation quality and flexibility.

Furthermore, some methods have proposed the idea of multi-layer Gaussian models. GSGAN and its variants~\cite{hyun2024gsgan, barthel2025cgsgan} introduced a 3D Generative Adversarial Network that generates a hierarchy of Gaussians, where finer-level Gaussians are parameterized by their coarser-level counterparts. However, these generative models do not achieve the disentanglement of identity and expression, making them difficult to transfer directly to the task of expression driving. Our proposed SplitAvatar introduces the autoregressive paradigm into the representation learning of 3DGS, effectively capturing complex expression details.

%% file: sec/3_method.tex
\vspace{-5pt}
\section{Method}
\label{sec:method}
\vspace{-5pt}
\subsection{Preliminary}
\vspace{-5pt}

\textbf{3DGS.} 3D Gaussian Splatting (3DGS) provides an efficient method for high-quality scene reconstruction that utilizes anisotropic 3D Gaussians. Each Gaussian is defined by its center (mean) $\mu$ and covariance matrix $\Sigma$:
\begin{equation}
    G(x)=e^{-\frac{1}{2}(x-\mu )^{T} \Sigma^{-1} (x-\mu )  } 
    \label{eq:3dgs}
\end{equation}
Furthermore, the covariance matrix $\Sigma$ is decomposed into a rotation matrix \(R\) and a scaling matrix \(S\) to ensure it is positive semi-definite, enabling the optimization process to proceed successfully:
\begin{equation}
  \Sigma  =RSS^{T} R^{T}  
  \label{eq:co}
\end{equation}
For rendering, each Gaussian is also assigned additional attributes, such as color $c$ and opacity $\alpha$. The color $C$ of a pixel is computed using a tile-based rasterizer by blending the Gaussians ordered by depth:
\begin{equation}
    C=\sum_{i=1}^{N}c_{i}\alpha _{i}^{'} \prod_{j=1}^{i-1} (1-\alpha _{j}^{'})
    \label{eq:render_gs}
\end{equation}
where $c_i$ is the color of each Gaussian. We utilize a neural network to predict $c_i$, which consists of 32-dimensional color features along with RGB information. We then use a neural renderer to generate the final image, which has been shown to yield better rendering detail~\cite{chu2024generalizable,li2023generalizable,chu2024gpavatar,ye2024real3d}. 

\vspace{-5pt}
\subsection{Overview}
\vspace{-5pt}

As illustrated in \cref{fig:struct}, our goal is to reconstruct a portrait from a source image while transferring the expression from one or more driving images onto the source person's face. We use a frozen Dinov2~\cite{oquab2023dinov2,darcet2023vision} model to extract facial features and feed them into the Gaussian feature adapter module to generate identity Gaussians. This module incorporates a trainable Vision Transformer (ViT) and adopts a dual-lift strategy similar to GAGAvatar~\cite{chu2024generalizable}. To capture the expression of driving image, we establish expression 3D Gaussians by a Gaussian generator based on the mesh vertices from the 3DMM estimator~\cite{li2017flame} to decouple facial expressions from identity. The generator comprises an MLP conditioned by Dinov2 features to preserve the identity of the source image. 

Specifically, the Gaussian generator produces one Gaussian at the position of each tracked mesh vertex, totaling 5023 Gaussians. This count is significantly smaller than the Gaussians produced by the Gaussian feature adapter. Subsequently, these expression-conditioned Gaussians are fed into an autoregressive structure. At each layer, a Gaussian splitting network (GSN) increases the number of Gaussians and capture multi-layer expression details. Concurrently, we employ a gating mechanism at each layer to dynamically prune redundant Gaussians.

\vspace{-5pt}
\subsection{Autoregressive Graph Network}
\label{subsec:GSN}
\vspace{-5pt}

\noindent\textbf{Gaussian Splitting Network.} We first define the expression Gaussians as a sequence of $L+1$ discrete layers $G_{exp}=( G_0,\dots ,G_{L} )$. The initial Gaussians from the Gaussian generator are set as $G_0$. We progressively predict Gaussian attributes for subsequent layers through the GSN to construct a Gaussian sequence. Each layer's Gaussian attributes depend on the preceding layer's Gaussians and latent features, where a single Gaussian can derive $k$ new Gaussians at the next layer. Each prediction step is referred to as a Gaussian splitting operation.

The core module of the GSN is a Graph Neural Network (GNN). Since the expression and pose estimated from the driving image inherently contains the topology of the FLAME model, we can convert the learning of facial details into information propagation among the nodes of the GNN. This allows Gaussians in key facial regions to receive more precise optimization signals. We utilize a Graph Attention Network (GAT)~\cite{velivckovic2018gat}, which allows the model to dynamically assign different importance weights to neighboring nodes. Specifically, the positions, opacities, rotations, scales, and color of $G_{i}$ are mapped to 256-dimensional feature $\mathcal{F}_{i}$ via a feature embedding module $\mathcal{E}$:
\begin{equation}
    \mathcal{F} _{i}=\mathcal{E} (G_{i})
\end{equation}
\(\mathcal{F}_{i}\) is inputted into the GSN to predict the Gaussian features of the next layer, and the actual Gaussian attributes $G_{i+1}$ are obtained via a Decoder $\mathcal{D}$:
\begin{equation}
    G_{i+1}=\lambda \mathcal{D} (N _{GSN}(\mathcal{F} _{i} ,E ^{i})+\mathcal{F} _{i})
\end{equation}
where $E^{i}$ denotes the edge information (adjacency matrix) for layer $i$. $\lambda$ is a scale factor of each attribute. By setting gradually decreasing values for opacity and scaling, different layers capture distinct levels of detail in the portrait. We also add a residual connection to the features to prevent gradient vanishing.


\noindent\textbf{Mesh topology extension for gaussian rigging.}
The Gaussian splitting operation causes the number of Gaussians to increase exponentially. This growth creates an inconsistency with the GNN's initial adjacency matrix, halting the feature learning process. To address this mismatch, we propose a mesh-topology splitting method that dynamically aligns the GNN's graph topology with the increased number of Gaussians.

The total edge set $E^{i}$ at layer $i$ consists of two parts: Topological Inheritance Edges $E^{i}_{\text{topo}}$ and Internal Connection Edges $E^{i}_{\text{conn}}$. The $E^{i}_{\text{topo}}$ preserve the original topology of $M^{0}$ (the FLAME model) within each of the $k$ mesh copies. Therefore, the number of vertices at layer $i$ is \(|V^{i}| = k^i  |V^{0}| = 5023 k^i\). The vertex ${V}^{i}$ at layer $i$ can be defined by two indices: its original graph index $j$ from $M^{0} = (V^{0}, E^{0})$ and its copy index $m \in \{1, \dots, k^{i}\}$:
\begin{equation}
    V^{i} = \left\{ \mathbf{v}_{j, m}^{i} \mid \mathbf{v}_j^{0} \in V^{0} \right\}
\end{equation}
\begin{equation}
    E^{i}_{\text{topo}} = \left\{ \left( \mathbf{v}_{j, m}^{i}, \mathbf{v}_{p, m}^{i} \right) \mid (\mathbf{v}_j^{0}, \mathbf{v}_p^{0}) \in E^{0} \right\}
\end{equation}

$E^{i}_{\text{conn}}$ forms a full connection among the $k^i$ new vertices that derived from the same parent vertex $\mathbf{v}_j^{0}$ to ensure information propagation among the split Gaussians.
\begin{equation}
    E^{i}_{\text{conn}} = \left\{ \left( \mathbf{v}_{j, m}^{i}, \mathbf{v}_{j, n}^{i} \right) \mid \mathbf{v}_j^{0} \in V^{0} \right\}
\end{equation}

where $n \in \{1, \dots, k^i\}$. The total edge set $E^{i}$ is the union of these two components:
\begin{equation}
    E^{i} = E^{i}_{\text{topo}} \cup E^{i}_{\text{conn}}
\end{equation}

This new graph $M^{i} = (V^{i}, E^{i})$ is then used as the input topology for the GNN at layer $i$.

\begin{figure*}[t]
    \centering
    \includegraphics[width=0.98\linewidth]{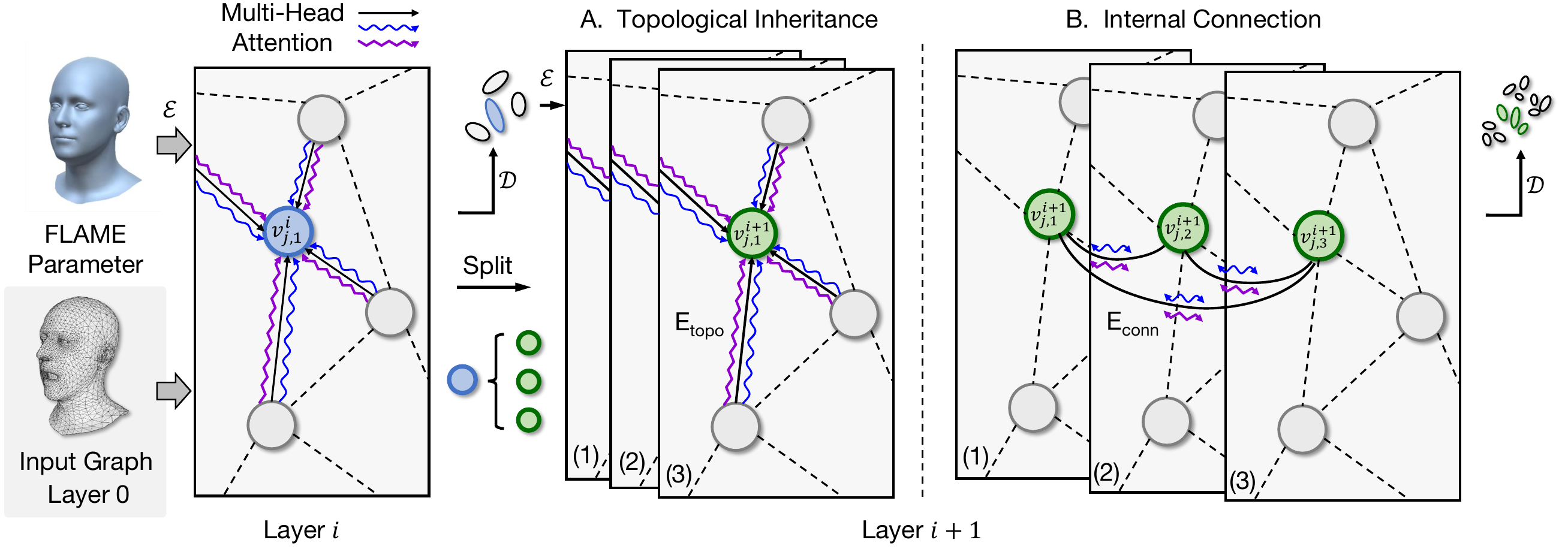}
    \vspace{-10pt}
    \caption{\textbf{Graph Splitting Network.} We embed the Gaussian attribute features from the previous layer into the GSN and predict Gaussian features through a GNN with an attention mechanism. The features are then decoded to obtain the split Gaussian attributes for the next layer. Given the graph information of the current layer, we define the edge topology of the next layer of GSN as two components: Topological Inheritance Edges $E^{i}_{\text{topo}}$ and Internal Connection Edges $E^{i}_{\text{conn}}$. The initial graph structure is derived from the FLAME mesh.}
    \vspace{-10pt}
    \label{fig:topo}
\end{figure*}
\vspace{-5pt}
\subsection{Gating Mechanism for 3D Gaussians}
\label{subsec:mask}
\vspace{-5pt}

Multiple splitting operations lead to an increase in Gaussian density, resulting in issues of redundancy and clustering. To achieve more flexible control over Gaussian density and distribution, we introduce a gating mechanism to dynamically filter out redundant Gaussians. 

\noindent\textbf{Soft Mask Network.}
Instead of using a fixed density control strategy~\cite{kerbl20233d,qian20243dgs}, we use a soft mask network $N_{mask}$ to predict a filter mask for the Gaussian Splats. Since the complexity of the human face varies across different regions, rigid filtering rules are insufficient for preserving all necessary rendering details. The soft mask $m_{i+1}$ for Gaussian $G_{i+1}$ is predicted based on the latent feature $\mathcal{F}_{i}$ from the previous layer:
\begin{equation}
    m_{i+1}=N_{mask}(\mathcal{F} _{i} )
\end{equation}

\noindent\textbf{Filtering Strategy.}
If non-compliant Gaussians are filtered out during each splitting operation, it would require recalculating the graph's edge indices. Subsequently, training the GNN with the retained Gaussians and new indices would incur significant overhead from the index computation. Furthermore, Gaussian features from different samples within the same batch may exhibit inconsistent lengths, potentially causing training failures. 
Therefore, we proposed a delayed filtering strategy. This strategy does not remove Gaussians during splitting phases; instead, it uses a mask to tag all the corresponding Gaussians, which are then simply ignored during the final rendering step. 

After each Gaussian splitting operations, the mask for layer $i$ is synchronously duplicated ${k}$ times, and then subjected to a logical AND operation with the output of $N_{mask}$ from layer $i+1$. The calculated result serves as the true mask for layer $i+1$. We first binarize the soft mask output using the quantizer function $q(x)$:
\begin{equation}
    q(x)=\begin{cases}
 1 & x>0.5 \\
 0 & \text{otherwise} 
\end{cases}
\end{equation}
The final logical mask $M_{i+1}^{\text{logical}}$ is then computed by combining the previous layer's mask $M_i$ with the newly predicted mask $m_{i+1}$:
\begin{equation}
    M_{i+1}^{logical}=q(m_{i+1})\wedge q(M_{i})
\end{equation}
\noindent\textbf{Gradient Propagation.}
The outputs from the GSN undergo two instances of gradient clipping due to the quantization function $q(\cdot)$ and the logical AND operation, which prevents gradients from propagating smoothly. To address this issue, we introduce the Straight-Through Estimator (STE) technique~\cite{courbariaux2015binaryconnect} within the relevant operations to ensure correct gradient backpropagation.
When calculating the gradient of the loss with respect to the quantization function $q(\cdot)$ using the chain rule, the STE ignores the true derivative (which is usually zero or undefined) and instead uses an identity function, setting the gradient to one.


The logical AND operation can be viewed as a variant of the quantization operation. During the forward pass, the AND truncates the mask to binary values; during the backward pass, the network retains only the gradient from the element-wise product of the two mask vectors, disregarding gradient propagation through the mask output after the logical AND truncation.

The operator $sg$ refers to a stop-gradient operation that allows forward computation and blocks backpropagation of gradients, similar to the method used in VQ-VAE~\cite{van2017vq-vae,razavi2019vq-vae-2}.

The mask update for gradient calculation is defined as:
\begin{equation}
    M_{i+1}=m_{i+1} M_{i} +sg(M_{i}^{logical}-m_{i+1} M_{i})
\end{equation}

\begin{figure*}[t]
    \centering
    \includegraphics[width=1\linewidth]{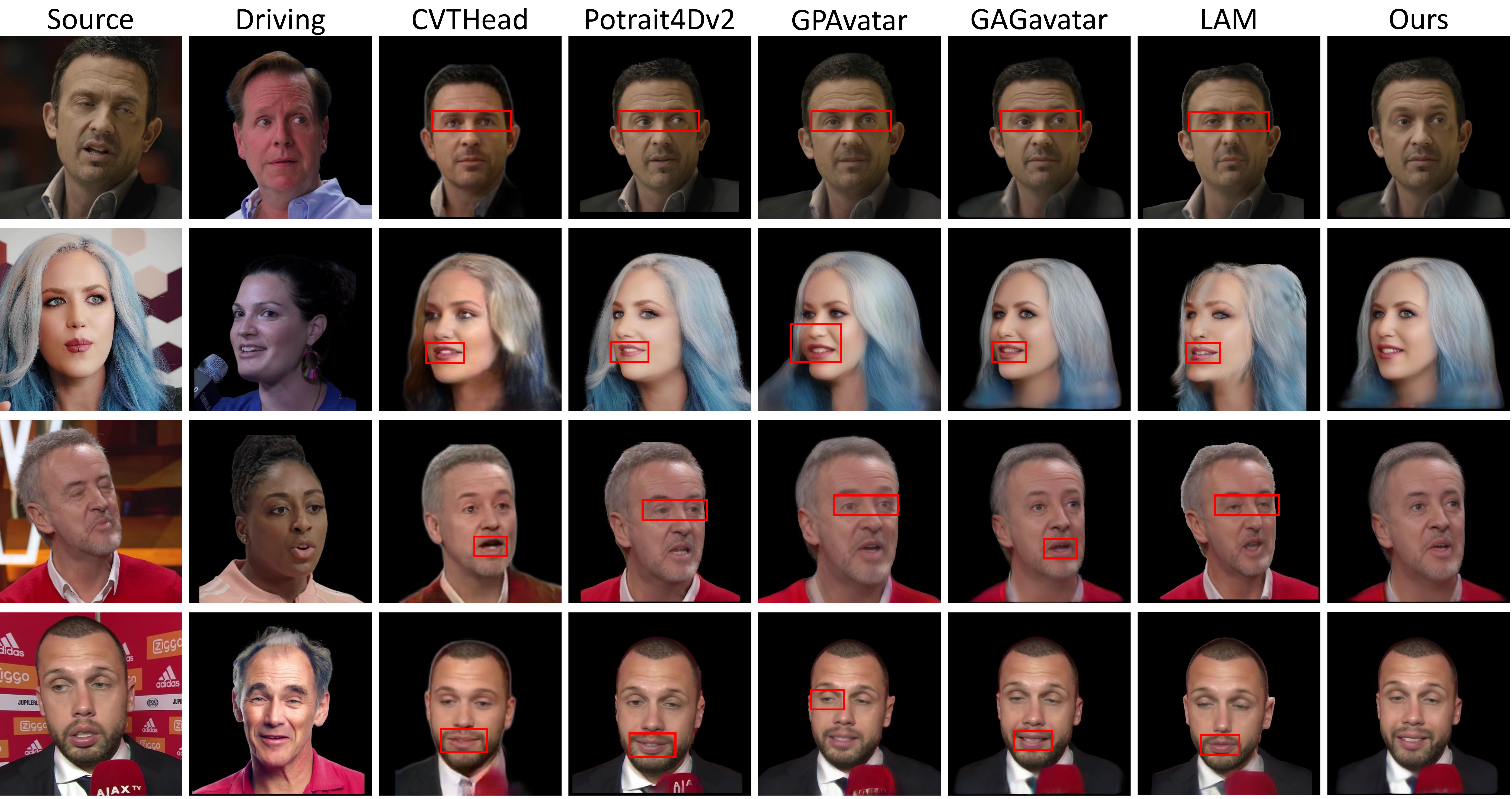}
    \vspace{-20pt}
    \caption{Cross-reenactment qualitative results on VFHQ datasets of different methods. Our method has more facial details.}
    \vspace{-5pt}
    \label{fig:result_vfhq}
\end{figure*}
\begin{table*}[t]
    \centering
    \caption{Experiment on VFHQ datasets. \colorbox{best}{Green} indicates the best and \colorbox{second}{yellow} indicates the second.}
    \vspace{-10pt}
    \resizebox{\textwidth}{!}{
    \begin{tabular}{l|ccccccc|ccc}
    \Xhline{1.2pt}
    \multirow{2}{*}{Method} & 
    \multicolumn{7}{c|}{Self Reenactment} & \multicolumn{3}{c}{Cross Reenactment} \\
          & PSNR$\uparrow$ & SSIM$\uparrow$ & LPIPS$\downarrow$ & CSIM$\uparrow$ & AED$\downarrow$ & APD$\downarrow$ & AKD$\downarrow$ & CSIM$\uparrow$ & AED$\downarrow$ & APD$\downarrow$ \\ 
         \hline
        CVTHead &
        17.6517	& 0.7445&0.2467&0.1851& 0.1876&0.240&8.0365& 0.0861&\cellcolor{second}0.2607&0.290 \\ 
        LAM &16.8278&0.6603&0.2696& 0.5793&0.1647&0.441&6.3958& 0.4724&0.2867&0.442 \\
        GPAvatar&
        18.4517&0.7575&0.2078&0.6268&0.1383&0.207&3.9999&0.4233&	\cellcolor{best}\textbf{0.2601}& 0.338 \\ 
        Portrait4D-v2&
        18.1148&0.7414&0.2060&0.7640&0.1205&0.261&7.3310&\cellcolor{best}\textbf{0.6210}& 0.2709&0.294 \\ 
        GAGAvatar&
\cellcolor{second}20.9694&\cellcolor{second}0.8105&\cellcolor{second}0.1524&\cellcolor{second}0.7947&\cellcolor{second}0.1091&\cellcolor{second}0.126&\cellcolor{second}3.3979&0.5688&
        0.2679&\cellcolor{second}0.280 \\ 
        \hline
        Ours &\cellcolor{best}\textbf{21.0958}& \cellcolor{best}\textbf{0.8124} & \cellcolor{best}\textbf{0.1503}& \cellcolor{best}\textbf{0.7975} & \cellcolor{best}\textbf{0.1084}& \cellcolor{best}\textbf{0.120}& \cellcolor{best}\textbf{3.3793}& \cellcolor{second}0.5896 &  0.2674 &\cellcolor{best} \textbf{0.278} \\
    \Xhline{1.2pt}
    \end{tabular}
    }
    \vspace{-10pt}
    \label{table:vfhq}
\end{table*}

The final rendered Gaussian is a summation of the reconstructed identity Gaussians ($G_{identity}$) and the masked expression Gaussians from each layer ($G_i$):
\begin{equation}
    G=G_{identity}+\sum_{i=0}^{L} M_i\odot G_i 
\end{equation}
\vspace{-5pt}
\subsection{Training Strategy}
\vspace{-5pt}
To measure the difference between target image and generated image, We employ L1 loss and perceptual loss to the coarse image $I_c$ and the fine image $I_f$. The image reconstruction loss $\mathcal{L}_{\text{image}}$ is defined as:
\begin{equation}
    \mathcal{L}_{image} = \left \| I_c-I_t \right \| +\left \| I_f-I_t \right \|
\end{equation}
The perceptual loss $\mathcal{L}_{\text{percep}}$ encourages similarity in feature space, where $\varphi(\cdot)$ denotes the feature extraction function~\cite{johnson2016perceptual,zhang2018perceptual,ye2024real3d,chu2024generalizable}:
\begin{equation}
    \mathcal{L}_{percep} = \left \| \varphi (I_c)-\varphi (I_t) \right \|+\left \| \varphi (I_f)-\varphi (I_t) \right \|
\end{equation}

We used the lifting loss $\mathcal{L}_{lifting}$ proposed in GAGAvatar~\cite{chu2024generalizable} to encourage the lifting points to be as close as possible to tracked 3DMM vertices. The distance is calculated using the L2 loss:
\begin{equation}
\begin{aligned}
\mathcal{L}_{lifting }&= \left \| P_{3dmm}-Q\right \| \\
Q &=\left \{ \textstyle\arg\min_{q \in G_{\text{identity}}}\left \| p-q \right \|\mid  p \in P_{3dmm}
  \right \}
\end{aligned}
\end{equation}
where $P_{3dmm}$ is the vertices of the tracked FLAME mesh and $argmin$ means finding the nearest point in $P_{3dmm}$ to each $G_{identity}$ position. Similarly, we propose a splitting Gaussian loss $\mathcal{L}_{split}$ to make the Gaussians generated by GSN to closely align with the tracked mesh.
\begin{equation}
\begin{aligned}
\mathcal{L}_{split}&= \left \| P_{3dmm}-Q\right \| \\
Q&=\left \{ \textstyle\arg\min_{q \in G_{\text{split}}}\left \| p-q \right \|\mid  p\in P_{3dmm}
  \right \}
\end{aligned}
\end{equation}
The overall training objective is as follows: 
\begin{equation}
    \mathcal{L} = \mathcal{L}_{image}
+\lambda _p\mathcal{L}_{percep} + \lambda _l\mathcal{L}_{lifting}+\lambda _s\mathcal{L}_{split}
\end{equation}
In our experiments, we set the weights as $\lambda_p=0.5$, $\lambda_l=1$, and $\lambda_s=0.5$.

%% file: sec/4_experiment.tex
\vspace{-5pt}
\section{Experiment}
\label{sec:experiment}
\vspace{-5pt}

\begin{table*}[t]
    \centering
    \caption{Experiment on HDTF datasets. \colorbox{best}{Green} indicates the best and \colorbox{second}{yellow} indicates the second.}
    \vspace{-10pt}
    \resizebox{\textwidth}{!}{
    \begin{tabular}{l|ccccccc|ccc}
    \Xhline{1.2pt}
    \multirow{2}{*}{Method} & 
    \multicolumn{7}{c|}{Self Reenactment} & \multicolumn{3}{c}{Cross Reenactment} \\
          & PSNR$\uparrow$ & SSIM$\uparrow$ & LPIPS$\downarrow$ & CSIM$\uparrow$ & AED$\downarrow$ & APD$\downarrow$ & AKD$\downarrow$ & CSIM$\uparrow$ & AED$\downarrow$ & APD$\downarrow$ \\ 
         \hline
        CVTHead &19.5501&0.8079&0.1910&0.3087 &0.1642&0.123&6.4926&0.2793 &\cellcolor{best}\textbf{0.2395}&\cellcolor{second} 0.153
         \\ 
        LAM &19.5718&0.7521&0.2008&0.6652&0.1761 & 0.159&6.4290& 0.6604& 0.2880&0.195
         \\
        GPAvatar&18.1250& 0.7932&0.1989& 0.8135& 0.1136&0.129&3.5576&0.7843&0.2749& 0.182
         \\ 
        Portrait4D-v2&19.6685&0.8039&0.1573&0.8424 &0.1071&0.157&6.0051&0.8442 &\cellcolor{second} 0.2612&0.177
         \\ 
        GAGAvatar&\cellcolor{second} 23.7415 & \cellcolor{second} 0.8778&\cellcolor{second} 0.1195& \cellcolor{second} 0.8642& \cellcolor{second} 0.1033& \cellcolor{second} 0.075 &\cellcolor{second} 3.1147&\cellcolor{second} 0.8457& 0.2709 &0.154
          \\  
        \hline
        Ours &\cellcolor{best}\textbf{24.1581}& \cellcolor{best}\textbf{0.8809} & \cellcolor{best}\textbf{0.1136}& \cellcolor{best}\textbf{0.8668} & \cellcolor{best}\textbf{ 0.1019 }& \cellcolor{best}\textbf{ 0.073 }& \cellcolor{best}\textbf{3.1052}& \cellcolor{best} \textbf{0.8547} &  0.2704  &\cellcolor{best} \textbf{0.152 } \\
    \Xhline{1.2pt}
    \end{tabular}
    }
    \vspace{-10pt}
    \label{table:hdtf}
\end{table*}
\subsection{Experiment setup}
\vspace{-5pt}

\begin{table*}[t]
    \centering
    \caption{Ablation study to validate the effectiveness of different components. We conducted experiments on the VFHQ dataset.}
    \vspace{-10pt}
    \resizebox{\textwidth}{!}{
    \begin{tabular}{l|ccccccc|ccc}
    \Xhline{1.2pt}
    \multirow{2}{*}{Method} & 
    \multicolumn{7}{c|}{Self Reenactment} & \multicolumn{3}{c}{Cross Reenactment} \\
          & PSNR$\uparrow$ & SSIM$\uparrow$ & LPIPS$\downarrow$ & CSIM$\uparrow$ & AED$\downarrow$ & APD$\downarrow$ & AKD$\downarrow$ & CSIM$\uparrow$ & AED$\downarrow$ & APD$\downarrow$ \\ 
         \hline
        Ours &\textbf{21.0958}& \textbf{0.8124} & \textbf{0.1503}&0.7975 & \textbf{0.1084}& \textbf{0.120}&\textbf{ 3.3793}& 0.5896 &\textbf{ 0.2674 }& 0.278 \\
        w/o AR $N_{GSN}$ &20.8219&0.8052& 0.1541 &\textbf{0.8018}&0.1424 &0.149&  4.1383& 0.6353 & 0.3137 & 0.299\\ 
        w/o GNN&21.0798&0.8103&0.1509&0.7985&0.1087&0.127&3.4119&0.6018& 0.2696& 0.284 \\ 
        w/o feature embedding &21.0524&0.8111&0.1512&0.7980& 0.1092&0.129&3.4096&0.5893&0.2694& \textbf{0.268} \\ 
        w/o $N_{mask}$ &20.7422&0.8040& 0.1616& 0.7943& 0.1481& 0.155& 4.2790 & \textbf{0.6408} & 0.3149 & 0.325 \\ 
    \Xhline{1.2pt}
    \end{tabular}
    }
    \vspace{-10pt}
    \label{table:ablation1}
\end{table*}
\begin{figure*}[ht]
    \centering
    \includegraphics[width=1\linewidth]{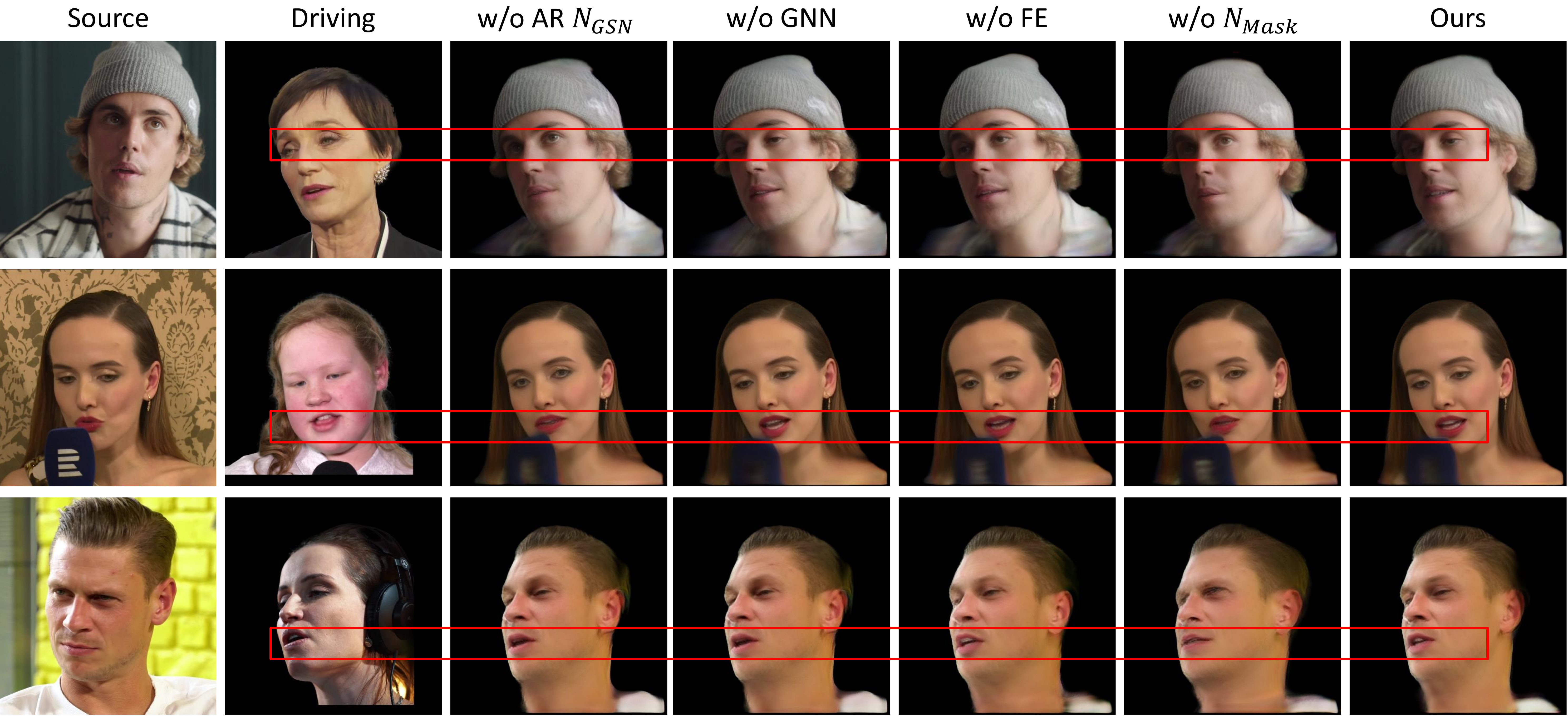}
    \vspace{-10pt}
    \caption{Ablation results on VFHQ datasets. w/o AR $N_{GSN}$ means replacing autoregressive GSN with a single layer and set the splitting factor $k$ to 1. w/o GNN means replacing the GNN with an MLP. FE denotes to feature embedding to positions, rotations, scales and opacities. w/o $N_{Mask}$ means removing the mask network at training and using all the predicted expression Gaussians. Our full method capture the most accurate facial expression from the driving image, as well as reconstructing the identity detail like teeth.}
    \label{fig:ablation}
    \vspace{-10pt}
\end{figure*}
\begin{table*}[t]
    \centering
    \caption{Ablation study on different splitting factor $k$ and layers $L$ setting.}
    \vspace{-10pt}
    \resizebox{\textwidth}{!}{
    \begin{tabular}{l|ccccccc|ccc}
    \Xhline{1.2pt}
    \multirow{2}{*}{\makecell{Method \\Split$\times$layer}} & 
    \multicolumn{7}{c}{Self Reenactment} & \multicolumn{3}{c}{Cross Reenactment} \\
          & PSNR$\uparrow$ & SSIM$\uparrow$ & LPIPS$\downarrow$ & CSIM$\uparrow$ & AED$\downarrow$ & APD$\downarrow$ & AKD$\downarrow$ & CSIM$\uparrow$ & AED$\downarrow$ & APD$\downarrow$ \\ 
         \hline
        $1 \times  1$ &20.8219&0.8052& 0.1541 &\textbf{0.8018}&0.1424 &0.149&  4.1383& 0.6353 & 0.3137 & 0.299\\ 
        $2\times 2$ &21.0531&\textbf{0.8130}&\textbf{0.1500}& 0.7980 &\textbf{0.1076}& 0.121 & 3.4020 & 0.5839&0.2682&0.272 \\ 
        $2\times 3$ (Ours) &\textbf{21.0958}&0.8124 & 0.1503& 0.7975 & 0.1084& \textbf{0.120}& \textbf{3.3793}& 0.5896 & \textbf{0.2674} & 0.278 \\
        $3\times 2$ &20.7876&0.8052&0.1550& 0.7938 &0.1448&0.147& 4.1529 & 0.6310 & 0.3136& 0.323\\ 
        $3\times 3$ &20.8274&0.8055&0.1555& 0.7967  & 0.1437&0.150& 4.1512 & \textbf{0.6403} & 0.3116& 0.303 \\ 
        $4\times 2$ &21.0657&0.8105&0.1501 & 0.7931 & \textbf{0.1076}&0.126 &3.4304 & 0.5795 & 0.2687& \textbf{0.271}\\ 
    \Xhline{1.2pt}
    \end{tabular}
    }
    \vspace{-10pt}
    \label{table:ablation2}
\end{table*}

\textbf{Datasets.}
We trained on the VFHQ dataset~\cite{xie2022vfhq} by sampling every 5th frame, keeping only images with detected faces. Our final training set contains 636,359 frames from 14,513 clips, including tracked camera pose and FLAME parameters. Following~\cite{chu2024generalizable}, we removed backgrounds and resized all images to $512 \times 512$.
For evaluation, we used 50 VFHQ test videos, designating the first frame as source and sampling 100 random frames as driving/target images. We also evaluated on the HDTF dataset~\cite{zhang2021hdtf}, using the same split as previous work~\cite{li2023generalizable,ma2023otavatar} and sampling 230 random frames per clip.

\noindent\textbf{Implementation Details.} We implemented our method using PyTorch and utilized the Adam optimizer~\cite{kingma2014adam}. The training learning rate was set to $1.0\text{e-}4$, and the batch size was set to 4. We trained the model for 200,000 iterations, which cumulatively consumed 77 GPU hours on a NVIDIA A100 GPU.

The Gaussian feature adapter comprises two trainable Vision Transformers~\cite{oquab2023dinov2}, both configured with $\text{depth}=12$ and $\text{heads}=8$. We extract features from the last layer of the intermediate outputs, which are then upsampled to $296 \times 296$ via transposed convolutions to predict the Gaussian parameters. Following~\cite{chu2024generalizable}, we do not directly predict the mean (position) of the Gaussians; instead, we predict a forward and backward offset relative to the feature plane produced by a frozen DINOv2~\cite{oquab2023dinov2} backbone. For the GSN, we employ a Graph Attention Network configured with $\text{heads}=4$, and utilize an MLP with a 256-dimensional feature input to predict the mask for each autoregressive layer. We set the number of autoregressive layers $L$ to 3, and the splitting factor $k$ to 2, which means two Gaussians are generated from one in each splitting operation. The inference speed reaches 41 FPS with a low GPU memory consumption of 3.1 GB, demonstrating the efficiency of our method.

\vspace{-5pt}
\subsection{Experiment results}
\vspace{-5pt}
\textbf{Qualitative results.} 
We conducted comparative experiments between several existing State-of-the-Art methods and our proposed approach. These methods include CVTHead~\cite{ma2024cvthead}, LAM~\cite{he2025lam}, GPAvatar~\cite{chu2024gpavatar}, Portrait4D-v2~\cite{deng2024portrait4d2}, and GAGavatar~\cite{chu2024generalizable}. To ensure a fair comparison, we used the official implementations of these methods to generate result images, then cropped and removed the background for all results before comparison.

~\cref{fig:result_vfhq} shows the qualitative results, demonstrating that our method reconstructs the human head avatar better than other approaches, particularly the rich details around facial features such as the eyes and an opened mouth with teeth.

\noindent\textbf{Quantitative results.}
We also performed a quantitative evaluation of our model. We assessed the performance of both self-reenactment and cross-reenactment on the VFHQ and HDTF datasets to test our method's ability to reconstruct the human portrait, preserve identity features, and transfer the driving expression.

For self-reenactment, we evaluated three widely recognized image quality metrics: PSNR, SSIM~\cite{wang2004ssim}, and LPIPS~\cite{zhang2018lpips} to quantify the similarity between the driving images and the synthesized images. Specifically, LPIPS computes perceptual differences using features extracted from a VGG network~\cite{simonyan2014vgg}. For identity similarity, we used the ArcFace model~\cite{deng2019arcface} to extract identity features from both the source and generated images, then calculated their cosine similarity (CSIM). We further used a 3DMM estimator~\cite{deng2019deep3d} to compute the average expression distance (AED) and average pose distance (APD), which measure the $L_1$ distance of expression and pose parameters between the driving and generated images. We also calculated the average keypoint distance (AKD) using a facial landmark detector~\cite{bulat2017alignment}, which extracts 68 facial keypoints from both images and computes the average Euclidean distance between corresponding normalized keypoints. For cross-reenactment, since there is no ground truth available, we followed previous methods~\cite{li2023generalizable,chu2024gpavatar,chu2024generalizable} and evaluated CSIM, APD, and AKD.

~\cref{table:vfhq} and ~\cref{table:hdtf} present the quantitative experimental results on the VFHQ and HDTF datasets, respectively. Our method outperforms previous methods in both reconstruction quality and expression similarity. However, the identity consistency in cross-reenactment is slightly lower than some competing methods. We attribute this to two factors: first, the 3DMM estimator we used does not fully decouple identity and expression; second, the increased network capacity tends to prioritize learning richer facial details at the expense of strict identity preservation. We consider this a reasonable trade-off, provided that the inference time is not significantly affected.

\vspace{-5pt}
\subsection{Ablation Study}
\vspace{-5pt}

\noindent\textbf{Effect of the autoregressive splitting.} We investigate the role of GSN in reconstructing the head avatar and enhancing the details of the driving expression, specifically the effect of the autoregressive structure for learning facial information. To validate its contribution, we replaced our autoregressive GSN with a single layer and set the splitting factor $k$ to 1 to examine whether its predicted Gaussian attributes could produce the same rendering quality. As shown in the second row of ~\cref{table:ablation1} (w/o AR $N_{\text{GSN}}$), we observe a slight increase in identity similarity (CSIM). This suggests that with limited model capacity or deformation ability, the model is compelled to focus more on preserving the original identity at the expense of accurately fitting complex expressions and poses. In contrast, the autoregressive structure enhances the reconstruction quality of avatars and improves the accuracy of facial expressions.

\noindent\textbf{Effect of the graph splitting network.} To validate the contribution of the GNN to information propagation across the 3DMM topology and expression space, we replaced the GSN module with a MLP of equivalent input and output dimensions. As shown in ~\cref{table:ablation1}, the reconstruction quality (PSNR, SSIM, LPIPS) and facial expression accuracy (AED, APD, AKD) decrease but identity similarity (CSIM) slightly increase when the GNN is replaced. This suggests that the simplified capacity of the MLP forces the model to prioritize identity preservation, sacrificing the ability to capture complex, high-frequency expression and pose details. The degradation in metrics like AKD and AED confirms that the GNN module is crucial for leveraging graph topology to enable accurate and detailed expression transfer.

\noindent\textbf{Effect of the Feature Embedding.} To verify whether mapping the Gaussian attributes to the latent feature space helps improve the quality of the synthesized images, we removed both the feature embedding $\mathcal{E}$ and the decoder $\mathcal{D}$ and directly fed the Gaussian attributes into $N_{GSN}$. The slight decrease in the metrics shown in the fourth row of ~\cref{table:ablation1} (w/o {$\text{feature embedding}$) indicates that the basic network structure is sufficient for core reconstruction task. ~\cref{fig:ablation} demonstrates that feature embedding helps capture relationships among Gaussian parameters, which provides richer information and improves generalization performance.

\noindent\textbf{Effect of the Mask Network.} To verify the effectiveness of the mask network in filtering Gaussians and its specific impact on rendering results, we removed the entire mask network during training and rendered the image directly using all predicted Gaussians. The cross-reenactment result in \cref{fig:ablation} (w/o ${N}_{\text{mask}}$) indicates that the model tends to render a nearly static face because it cannot accurately perform filtering and reenactment. As shown in the fifth row of ~\cref{table:ablation1}, although the CSIM in cross-reenactment is notably higher, this is due to identity leakage rather than genuine identity preservation. Without the mask network, the model tends to retain the source appearance instead of transferring expressions, leading to degraded motion-related metrics (\ie, AED and APD). In contrast, our full model achieves a well-balanced trade-off between identity preservation and motion accuracy, delivering superior expression transfer while maintaining competitive identity similarity.

\noindent\textbf{Effect of Different Splitting Factors and Layers.} We further explored the effects of setting different Gaussian splitting factor $k$ and layers $L$ for the autoregressive process. The experimental results are shown in ~\cref{table:ablation2}. The $1 \times 1$ configuration performs well on CSIM (identity similarity) but is relatively poor in reconstruction quality and reenactment accuracy. This indicates that when information is limited, our model tends to prioritize identity preservation. When splitting parameters are set to medium settings ($2 \times L$ and $4 \times 2$), these configurations demonstrate significant advantages in both reconstruction quality and reenactment accuracy. This suggests that a moderate number of splitting layers ($L=2$ or $3$) and splitting factor $k=2$ effectively increase the model capacity and detail expression capability. The increase in model capacity enables it to learn facial expression and pose variations better, at the cost of slightly compromising identity preservation. Further increase in splitting count and layers did not lead to significant performance gains and even showed degradation in some metrics ($3 \times 3$ setting in ~\cref{table:ablation2}). This suggests that an excessive number of Gaussians results in overfitting or information redundancy, thereby impairing the model's generalization and expressiveness.


%% file: sec/5_conclusion.tex
\vspace{-5pt}
\section{Conclusion}
\vspace{-5pt}
\label{sec:conclusion}
We propose SplitAvatar, a novel method for single-image human portrait reconstruction and expression driving. Our key idea is to utilize an autoregressive GNN to split the limited number of Gaussians obtained from a 3DMM estimator, which learns multi-level and richer expression details.
Through different configurations, the number of Gaussians can be manually controlled to match any vision-based methods, providing high controllability and portability. We also introduced a graph topology expansion method and a gating mechanism with a soft mask network to optimize the density and distribution of Gaussians, achieving superior driving effects. Experimental results demonstrate that our autoregressive structure and graph splitting network effectively increases Gaussians amount while capturing human facial features by GNN-guided Gaussians. Furthermore, the mask network enables dynamic pruning, removing redundant Gaussians caused by the splitting operation.


%% file: sec/6_suppl.tex
\section{More Details on Experiments}
\subsection{Implementation}
The Gaussian feature adapter comprises two trainable Vision Transformers~\cite{oquab2023dinov2}, both configured with $\text{depth}=12$ and $\text{heads}=8$. We extract features from the last layer of the intermediate outputs. The resulting features are then upsampled to $296 \times 296$ via transposed convolutions, subsequently predicting the Gaussian parameters. Our strategy, similar to ~\cite{chu2024generalizable}, avoids directly predicting the mean (position) of the Gaussians. Instead, we predict a forward and backward offset relative to the feature plane predicted by a frozen DINOv2~\cite{oquab2023dinov2} backbone.

Regarding the GSN, we employ a Graph Attention Network configured with $\text{heads}=4$. Furthermore, we utilize an MLP with a 256-dimensional feature input to predict the mask for each autoregressive layer.

\subsection{Metrics}
We compare our method with other works in the following three aspects: (1) Quality of synthesized images in the self reenactment task. (2) Identity similarity with source image. (3) Facial expression accuracy. 

\noindent\textbf{Quality of synthesized images.} We utilize three widely recognized image quality metrics \textbf{PSNR}, \textbf{SSIM}~\cite{wang2004ssim}, and \textbf{LPIPS}~\cite{zhang2018lpips} to quantify the similarity between the driving images and the synthesized images. Specifically, LPIPS computes perceptual differences using features extracted from a VGG network~\cite{simonyan2014vgg}.

\noindent\textbf{Identity similarity.}
We employ the pre-trained ArcFace model \cite{deng2019arcface} to extract high-dimensional identity features from both the source image and the generated image. The \textbf{cosine similarity (CSIM)} is then calculated as:

\begin{equation}
    \text{CSIM} = \frac{A \cdot B}{\|A\| \|B\|}
\end{equation}

where $A$ is the normalized identity feature vector of the source image and $B$ is the normalized identity feature vector of the generated image.

\noindent\textbf{Facial expression accuracy.} We utilized three metrics:

\textbf{Average Expression Distance (AED)} and \textbf{Average Pose Distance (APD) :} These metrics compute the $L_1$ loss of the expression and pose parameters, respectively, extracted by a 3DMM estimator\cite{deng2019deep3d} between the driving images and the result images.

\textbf{Average Keypoint Distance (AKD):} We use a facial landmark detector\cite{bulat2017alignment} to extract 68 facial landmark keypoints from the generated image and the driving image, then calculate the average Euclidean distance between the corresponding normalized keypoints.

\section{More Experimental Results}

\begin{figure*}
    \centering
    \includegraphics[width=1\linewidth]{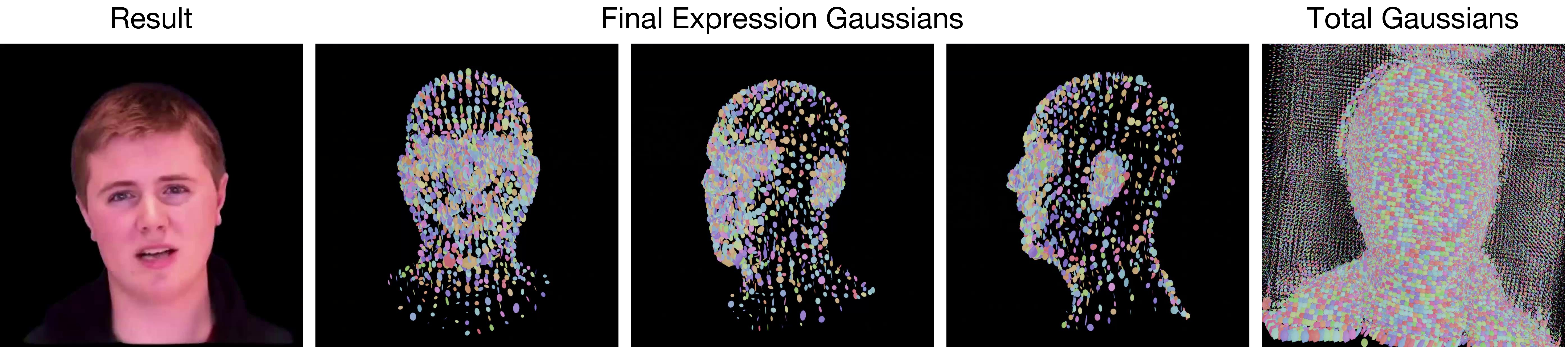}
    \caption{Gaussians visualization of our method. Results means the final image after neural rendering. Final Expression Gaussians present multi-view distribution of all masked expression Gaussians $G_{exp}$. Total Gaussians visualized both identity Gaussians and expression Gaussians. The Gaussian distribution is concentrated around the facial features, indicating that our method dynamically controls the density of the expression Gaussian.}
    \label{fig:gaussian}
\end{figure*}

\subsection{Visual Results of Gaussians}
We visualize the preserved Gaussians using Blender. As shown in ~\cref{fig:gaussian}, the final expression Gaussians distribution is concentrated around the facial features, such as the eyes, mouth and ears. This demonstrates that our autoregressive GSN and mask network successfully capture fine-grained expression details while dynamically controlling Gaussian density. The background areas tend to become very small in total Gaussians, which indicates that identity Gaussians have successfully reconstructed the human figure. 

\begin{table*}[!t]
    \centering
    \caption{Comparasion with baseline expanded expression output by $15\times$. }
    \vspace{-10pt}
    \resizebox{\textwidth}{!}{
    \begin{tabular}{l|ccccccc|ccc}
    \Xhline{1.2pt}
    \multirow{2}{*}{Method} & 
    \multicolumn{7}{c|}{Self Reenactment} & \multicolumn{3}{c}{Cross Reenactment} \\
          & PSNR$\uparrow$ & SSIM$\uparrow$ & LPIPS$\downarrow$ & CSIM$\uparrow$ & AED$\downarrow$ & APD$\downarrow$ & AKD$\downarrow$ & CSIM$\uparrow$ & AED$\downarrow$ & APD$\downarrow$ \\ 
        \hline
        Ours &\textbf{21.0958}& \textbf{0.8124} & \textbf{0.1503}& \textbf{0.7975} & \textbf{0.1084}& \textbf{0.120}& \textbf{3.3793}& 0.5896 &  \textbf{0.2674} &\textbf{0.278} \\
        Baseline $15\times$ &20.9251&0.8016&0.1588&0.7928&0.1450&0.153&4.1602&\textbf{0.6390}& 0.3114& 0.312 \\ 
    \Xhline{1.2pt}
    \end{tabular}
    }
    \vspace{-10pt}
    \label{table:ablation_gnn_metric}
\end{table*}

\subsection{Comparison with baseline using increased Gaussians}

To further validate the efficiency and effectiveness of our autoregressive splitting strategy, we compare with a variant of baseline that expands the expression output by $15\times$, matching the total number of Gaussians produced by our method. Both models are trained under identical configurations, including the same training steps, learning rate, batch size, and dataset, ensuring a fair and controlled comparison.

As shown in ~\cref{table:ablation_gnn_metric}, our method outperforms baseline $15\times$ across nearly all metrics in both self-reenactment and cross-reenactment, demonstrating superior reconstruction quality and expression transfer accuracy. Notably, baseline $15\times$ exhibits a higher CSIM in cross-reenactment. As discussed in ablation study, it is attributed to its less accurate expression transfer: when the driving motion is not faithfully reproduced, the generated face naturally retains more similarity to the source identity, resulting in an inflated CSIM that does not reflect genuine identity preservation capability.